%% file: smartmask.tex
\documentclass[10pt,twocolumn,letterpaper]{article}

\usepackage[pagenumbers]{cvpr} 

\input{preamble}

\usepackage{graphicx}
\graphicspath{{assets/}} 
\usepackage{amsmath}
\usepackage{amssymb}
\usepackage{booktabs}
\usepackage{enumitem}
\usepackage{algorithm}
\usepackage{algpseudocode}
\usepackage{bm}
\usepackage{multirow}
\usepackage{caption}
\usepackage{subcaption}
\usepackage{tikz}

\usepackage{pifont}
\usepackage[pagebackref,breaklinks,colorlinks]{hyperref}

\newcommand{\bline}[1]{\textcolor{orange}{\emph{#1}}}

\makeatletter
\def \smask{\emph{smartmask} }  \def \Smask{\emph{SmartMask} } 
\makeatother

\definecolor{cvprblue}{rgb}{0.21,0.49,0.74}

\usepackage[capitalize]{cleveref}
\crefname{section}{Sec.}{Secs.}
\Crefname{section}{Section}{Sections}
\Crefname{table}{Table}{Tables}
\crefname{table}{Tab.}{Tabs.}

\def\paperID{1050} 
\def\confName{CVPR}
\def\confYear{2024}

\begin{document}

\title{\textcolor{cvprblue}{SmartMask:} Context Aware High-Fidelity Mask Generation for \\Fine-grained Object Insertion and Layout Control}


\author{Jaskirat Singh$^{1,2}$ \quad Jianming Zhang$^1$ \quad Qing Liu$^1$ \quad Cameron Smith$^1$ \quad  Zhe Lin$^1$ \quad  Liang Zheng$^2$\\
$^1$Adobe Research \qquad $^2$Australian National University\\
{\url{https://smartmask-gen.github.io}}
}


\twocolumn[{
\maketitle
\begin{center}
    \vskip -0.25in
    \captionsetup{type=figure}
    \begin{subfigure}[b]{0.95\textwidth}
         \centering
         \includegraphics[width=\textwidth]{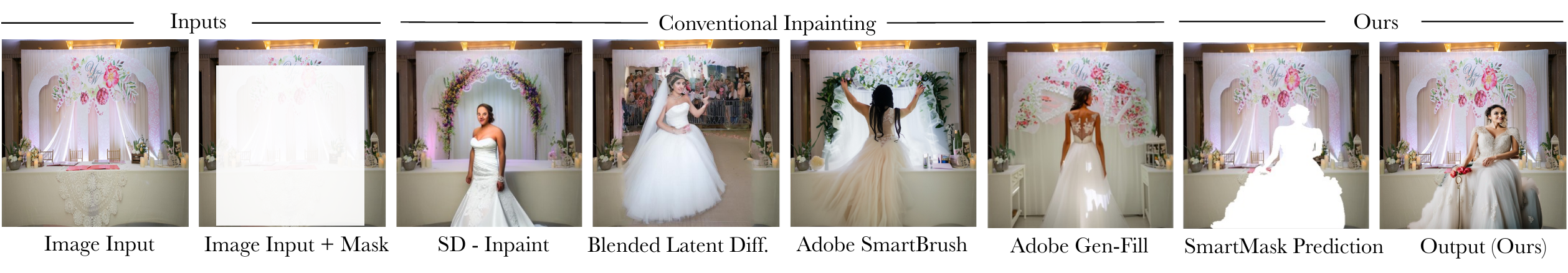}
         \vskip -0.05in
         \caption{SmartMask for performing object insertion (``bride'') with better background preservation.}
     \end{subfigure} 
     \vskip 0.05in
     \begin{subfigure}[b]{0.95\textwidth}
         \centering
         \includegraphics[width=\textwidth]{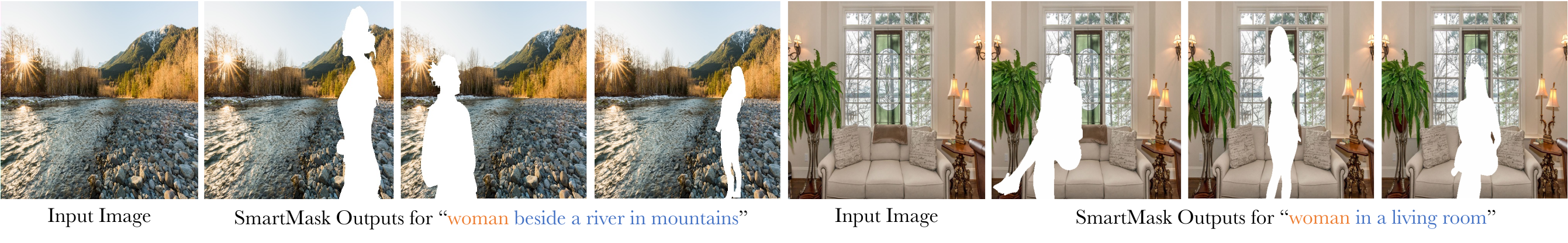}
         \vskip -0.05in
         \caption{SmartMask for mask-free object insertion where it provides diverse suggestions without user bounding box input.}
     \end{subfigure} 
     \begin{subfigure}[b]{0.95\textwidth}
         \centering
         \includegraphics[width=\textwidth]{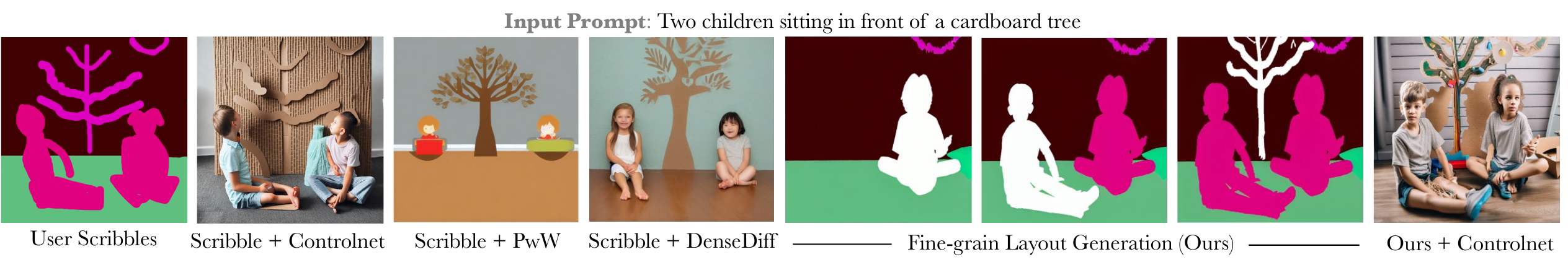}
         \vskip -0.02in
         \caption{SmartMask for fine-grain layout design for layout to image generation.}
     \end{subfigure} 
    \vskip -0.1in
    \caption{\emph{\textbf{Overview}}. 
    We introduce \Smask which allows any novice user to generate high-fidelity masks for fine-grained object insertion and layout control. The proposed approach can be used for object insertion (a-b) where it not only allows for image inpainting with better background preservation (a) but also provides diverse suggestions for mask-free object insertion at different positions and scales. We also find that when used iteratively \Smask can be used for fine-grained layout design (c) for better quality semantic-to-image generation.}
     \label{fig:overview}
\end{center}
}
]







\begin{abstract} 
    \vskip -0.15in
    The field of generative image inpainting and object insertion has made significant progress with the recent advent of latent diffusion models. Utilizing a precise object mask can greatly enhance these applications. However, due to the challenges users encounter in creating high-fidelity masks, there is a tendency for these methods to rely on more coarse masks (e.g., bounding box) for these applications. This results in limited control and compromised background content preservation. To overcome these limitations, we introduce SmartMask, which allows any novice user to create detailed masks for precise object insertion. 
    Combined with a ControlNet-Inpaint model, our experiments demonstrate that SmartMask achieves superior object insertion performance, preserving the background content more effectively than previous methods. 
    Notably, unlike prior works the proposed approach can also be used without user-mask guidance, which allows it to perform mask-free object insertion at diverse positions and scales.
    Furthermore, we find that when used iteratively  with a novel instruction-tuning based planning model, SmartMask can be used to design detailed layouts from scratch. As compared with user-scribble based layout design, we observe that SmartMask allows for better quality outputs with layout-to-image generation methods.
\vskip -0.2in
\end{abstract}

\section{Introduction}
\label{sec:intro}

Multi-modal object inpainting and insertion has gained widespread public attention with the recent advent of large-scale language-image (LLI) models \cite{avrahami2022blended,avrahami2023blended,nichol2021glide,xie2023smartbrush,yang2023uni,li2023gligen,rombach2021highresolution,podell2023sdxl,adobe2023firefly}. A novice user can gain significant control over the inserted object details by combining text-based conditioning with additional guidance from a coarse bounding-box or user-scribble mask. The text prompt can be used to describe the object semantics, while the coarse-mask provides control over the position and scale of the generated object.

While convenient, the use of a coarse-mask for object insertion suffers from two main limitations. \textbf{1) First,} the use of a coarse mask can often be undesirable as it tends to also modify the background regions surrounding the inserted object \cite{xie2023smartbrush, rombach2021highresolution} (refer Fig.~\ref{fig:overview}a). In order to minimize the background artifacts, recent works \cite{xie2023smartbrush} also explore the use of user-scribble based free-form mask instead of a bounding-box input. However, while feasible for describing coarse objects (\eg, mountains, teddy bear \etc), the generation of accurate free-form masks for objects with a number of fine-grain features (\eg, humans) can be quite challenging especially when limited to coarse user-scribbles. \textbf{2)} Furthermore, generating variations in position and scale of the inserted object can also be troublesome as it requires the user to provide a new scene-aware free-form mask to satisfy the geometric constraints at the new scene location.

To address these drawbacks, we introduce \emph{SmartMask}, a  context-aware diffusion model which allows a novice user to directly generate fine-grained mask suggestions for precise object insertion.
In particular, given an input object description and the overall scene context, \Smask generates scene-aware precise object masks which can then be used as input to a ControlNet-inpaint model \cite{zhang2023adding,diffusers} to perform object insertion while preserving the contents of the background image. As compared with coarse-mask based inpainting methods, we find that \Smask provides a highly convenient and controllable method for object insertion and can be used  1) \emph{with user-inputs (bounding box, scribbles etc.)}: where the user can specify location and shape for the target object , or 2) \emph{in a mask-free manner:}  where the model automatically generates diverse suggestions for object insertion at diverse positions and scales.

In addition to object insertion, we also find that SmartMask can be used for fine-grain layout design.  Existing segmentation-to-image (S2I) generation methods (\eg ControlNet \cite{zhang2023adding}) enable the generation of controllable image outputs from user-scribble based semantic segmentation maps.
However, generating a good quality semantic layout can be quite challenging if the user wants to generate a scene with objects that require fine-grain details for best description (\eg, humans, chairs \etc).
To address this challenge, we show that \Smask when used with a novel instruction-tuning \cite{touvron2023llama} based planning model allows the user to 
iteratively generate the desired scene layout from scratch.
As compared with scribble based layout generation, we find that the proposed approach allows the
users to better leverage existing S2I generation methods (\eg ControlNet \cite{zhang2023adding}) for higher quality output generation.

The main contributions of the paper are: 1) We propose \Smask which allows any novice user to generate precise object masks for finegrained object insertion with better background preservation.  2) We show that unlike prior works, the proposed approach can also be used for mask-free object insertion.  3) Finally, we demonstrate that \Smask can be used iteratively to generate detailed semantic layouts, which allows users to better leverage existing S2I generation methods for higher quality output generation.

\section{Related Work}
\label{sec:related-work}

\begin{figure*}[h!]
\vskip -0.15in
\begin{center}
\centerline{\includegraphics[width=1.\linewidth]{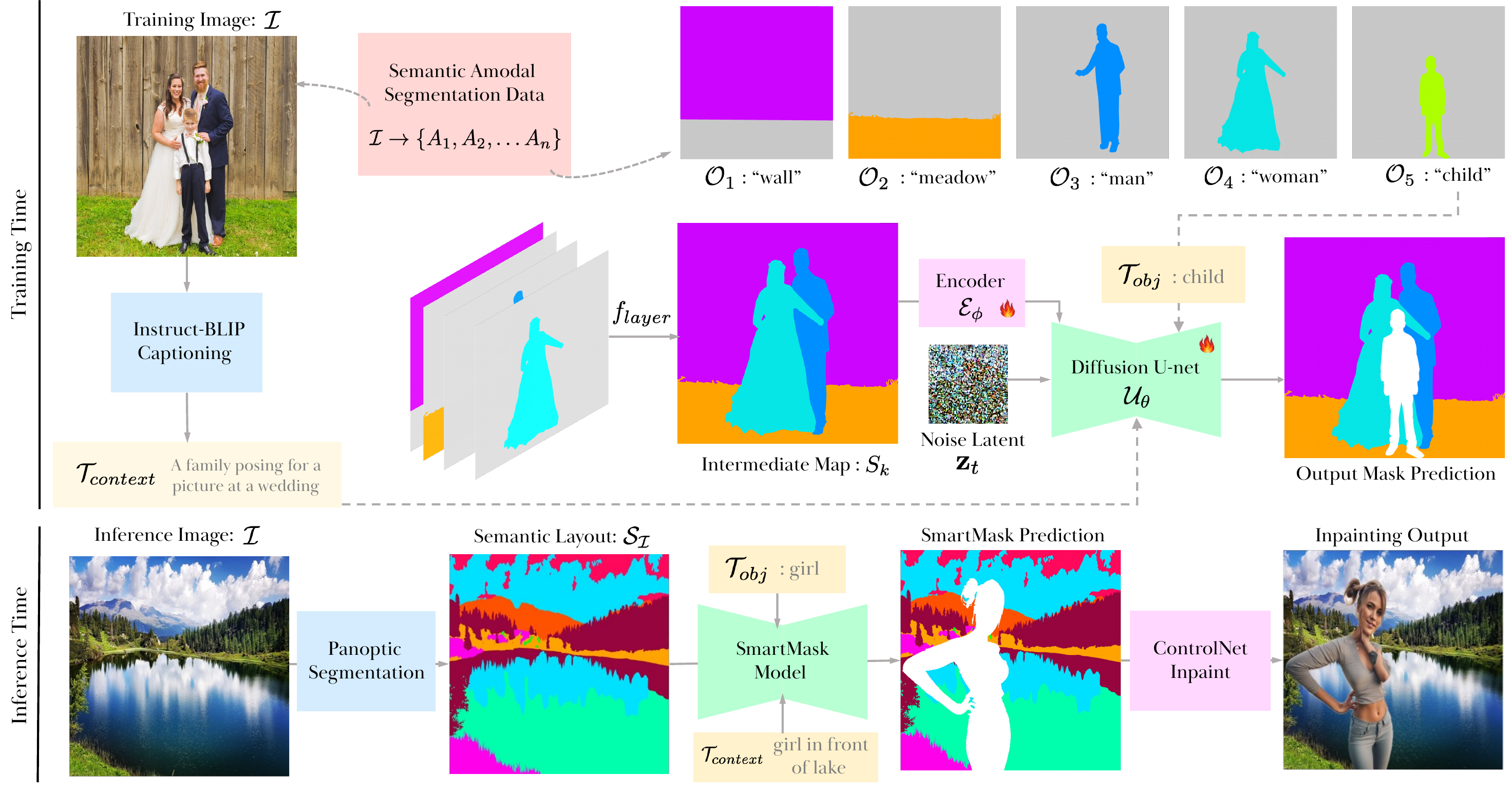}}
\vskip -0.1in
\caption{\emph{\textbf{Method Overview.}} A key idea behind \Smask is to leverage semantic amodal segmentation data \cite{zhu2017semantic,qi2019amodal} in order to obtain high-quality paired training annotations for mask-free (single or multi-step) object insertion. During training \emph{(top)}, given a training image $\mathcal{I}$ with caption $C$, we stack $k$ ordered instance maps $\{A_1,A_2,\dots A_k\}$ to obtain an intermediate semantic map $S_k$. The diffusion model is then trained to predict the instance map $A_{k+1}$, conditional on the semantic map $S_k$, $\mathcal{T}_{obj}\leftarrow \mathcal{O}_{k+1}$ and scene context $\mathcal{T}_{context} \leftarrow C$. During inference \emph{(bottom)}, given a real image $\mathcal{I}$, we first use a panoptic segmentation model to compute semantic map $\mathcal{S}_{I}$. The generated semantic layout is then directly used as input to the
trained diffusion model in order to predict the fine-grained
mask for the inserted object.}
\label{fig:method-overview}
\end{center}
\vskip -0.4in
\end{figure*}

\textbf{Diffusion based multi-modal image inpainting.} 
Recently, diffusion based multi-modal inpainting \cite{avrahami2022blended,avrahami2023blended,nichol2021glide,xie2023smartbrush,yang2023uni,li2023gligen,rombach2021highresolution,podell2023sdxl,adobe2023firefly} has gained widespread attention with advent of text-conditioned diffusion models \cite{rombach2021highresolution,nichol2021glide,saharia2022photorealistic,ramesh2022hierarchical,yu2022scaling}.
Despite their efficacy, prior works use coarse bounding box or user-scribble based masks for object inpainting which leads to poor background preservation around the inserted object. In contrast, \Smask directly allows user to generate precise masks for the target object, which can then be combined with a ControlNet-Inpaint \cite{zhang2023adding,rombach2021highresolution} to perform object insertion while better preserving the background contents.

\textbf{Mask-free object placement} has been studied in the context of image compositing methods \cite{zhu2023topnet,ghosh2019interactive,zhang2020learning, tripathi2019learning, lin2018st,niu2022fast,liu2021opa,lee2018context,zhou2022learning}, where given a cropped RGB object instance and a target image, the goal is to suggest different positions for the target object. In contrast, we study the problem of mask-free object insertion using text-only guidance. 

\textbf{Semantic-layout to image generation} methods have been explored to enable controllable image
synthesis from user-scribble based semantic segmentation
maps \cite{park2019semantic,zhu2020sean,lee2020maskgan, esser2021taming, isola2017image,liu2019learning, sushko2020you}.
Recently, \cite{balaji2022ediffi,singh2022high,densediffusion,chen2023training} propose a cross-attention based training-free approach for controlling the overall scene layout from coarse user-scribbles using text-conditioned diffusion models. Zhang \etal \cite{zhang2023adding} propose a versatile ControlNet model which allows the users to control the output layout on a more fine-grained level through an input semantic map. 
While effective, generating desired semantic layout with scribbles can itself be quite challenging for scenes with objects that require fine-grain details for best description (\eg, humans).
\Smask helps address this problem by allowing users to generate more fine-grained layouts  to facilitate better quality S2I generation.

\textbf{Bounding-box based layout generation}. The layout generation ability of the proposed approach can also be contrasted with bounding-box based specialized layout creation methods \cite{qu2023layoutllm,feng2023layoutgpt,zheng2019content,arroyo2021variational, gupta2021layouttransformer}. However, the generated layouts are represented only by coarse bbox locations. In contrast, the proposed approach allows the user to control the scene layouts on a more finegrain level including object shape / structure, occlusion relationships, location \etc. Furthermore, as illustrated in Fig.~\textcolor{red}{7}b, we note that \Smask generated layouts are highly controllable and allow for a range of custom operations such as adding, removing, modifying or moving objects through simple layer manipulations. 

\section{Our Method}
\label{sec:method}

Given an input image $\mathcal{I}$, object description $\mathcal{T}_{obj}$ and a textual description $\mathcal{T}_{context}$ describing the final scene context,
our goal is to predict a fine-grained mask $\mathcal{M}_{obj}$ for the target object. The object mask $\mathcal{M}_{obj}$ could then be used as input to a ControlNet-Inpaint model for fine-grained object insertion (Sec.~\ref{sec:inpaint-results}) or used to design detailed semantic layouts from scratch (Sec.~\ref{sec:layout-design-results}).
For instance, \Smask could be used for single object insertion,
where given an image  $\mathcal{I}$ depicting a man on a bench, $\{\mathcal{T}_{obj}:\text{\emph{`woman'}}\}$ and $\{\mathcal{T}_{context}:\text{\emph{`a couple sitting on a bench'}}\}$, our goal is to predict fine-grain binary mask $\mathcal{M}_{woman}$ which places the \emph{`woman'} in a manner such that the resulting scene aligns with overall scene context of a \emph{`a couple sitting on a bench'}. 

Similarly, \Smask could also be used for multiple object insertion. For instance, given an image $\mathcal{I}$ depicting a wedding, the user may wish to add multiple objects \{\emph{`man'}, \emph{`woman'} and \emph{`kid'}\} such that the final scene aligns with the $\{\mathcal{T}_{context}:\text{\emph{`a family posing for a picture at a wedding'}}\}$. Unlike prior image inpainting methods which are limited independently adding each object to the scene, 
the goal of the \smask model is to add each object in a manner such the generated \emph{`man'}, \emph{`woman'} and \emph{`kid'} appear to be \emph{`a family posing for a picture at a wedding'}. 
(refer Fig.~\ref{fig:method-overview}a, Sec.~\ref{sec:multi-object-insertion}). 

In the next sections, we describe the key \Smask components in detail. In particular, in Sec.~\ref{sec:seq-layer-pred} we discuss how \Smask can leverage semantic amodal segmentation data in order to obtain high-quality paired annotations for mask-free 
object insertion. We then discuss a simple data-augmentation strategy which allows the user to also control the position, shape of the inserted object using coarse inputs (bounding-box, scribbles) in Sec.~\ref{sec:data-aug}. Finally, in Sec.~\ref{sec:global-planning-model} we propose a visual-instruction tuning \cite{liu2023visual} based planning model which when used with \Smask, allows for generation of detailed semantic layouts from scratch.

\subsection{SmartMask for Mask-Free Object Insertion}
\label{sec:seq-layer-pred}

\textbf{Semantic-Space Task Formulation.} Directly learning a model for our task in \emph{pixel} space can be quite challenging, as it would require large-scale collection of training data for single or multi-step object insertion while maintaining background preservation. To address this key challenge, a core idea of our approach is to propose an equivalent task formulation which allows us to leverage large-scale semantic amodal segmentation data \cite{zhu2017semantic,qi2019amodal} for generating high-quality paired training data in the \emph{semantic} space
(Fig.~\ref{fig:method-overview}).

\textbf{SmartMask Training.}  In particular during training, given an image $\mathcal{I}$, with a sequence of ordered amodal semantic instance maps $\{A_1, A_2 \dots A_n\}$ and corresponding semantic object labels $\{\mathcal{O}_1, \mathcal{O}_2 \dots \mathcal{O}_n\}$, we first compute an intermediate layer semantic map as,
\begin{align}
    S_k = f_{layer}(\{A_1, A_2 \dots A_k\}) \quad where \ k \in [1,n]. \label{eq:intermediate-layer}
\end{align}
where $k$ is randomly chosen from $[1,n]$ and $f_{layer}$ is a layering operation which stacks the amodal semantic segmentation maps from $i \in [1,k]$ in an ordered manner (Fig.~\ref{fig:method-overview}).

We next train a diffusion-based mask prediction model $\mathcal{D}_{\theta}$ which takes as input the above computed intermediate semantic layer map $S_k$, textual description for next object $\mathcal{T}_{obj} \leftarrow \mathcal{O}_{k+1}$,  overall caption $\mathcal{T}_{context} \leftarrow C_\mathcal{I}$ (for image $\mathcal{I}$) and learns to predict the binary mask $\{A_{k+1}\}$ for the next object. To this end, we first pass the intermediate semantic map $S_k$ through a learnable encoder $\mathcal{E}_{\phi}$ to obtain the encoded features $\mathcal{E}_{\phi}(S_k)$. At any timestep $t$ of the reverse diffusion process, the denoising noise prediction $\epsilon_t$ is then computed conditional jointly on previous noise latent $\mathbf{z}_{t}$,  and model inputs \{$\mathcal{E}_{\theta}(S_k)$,$\mathcal{T}_{obj}$,  $\mathcal{T}_{context}$\}  as,
\begin{align}
    \Tilde{\mathbf{\epsilon}}_{pred} (t) = \mathcal{U}_{\theta} (\mathbf{z}_{t},\mathcal{E}_{\phi}(S_k),\mathcal{T}_{obj}, \mathcal{T}_{context}, t),
\end{align}
where $\mathcal{U}_{\theta}$ represents the U-Net of the diffusion model $\mathcal{D}$.

The overall diffusion model $\mathcal{D}$ is then trained to predict the next layer $A_{k+1}$ using the following diffusion loss,
\begin{align}
    \mathcal{L}_{t}(\theta, \phi) = \mathbf{E}_{t\sim[1,T], S_k, A_{k+1}, \epsilon_t}[\Vert \mathbf{\epsilon}_t - \Tilde{\mathbf{\epsilon}}_{pred} (t)\Vert^2],
\end{align}
where $T$ is total number of reverse diffusion steps, $\epsilon_t \sim \mathcal{N}(0,I)$ is sampled from a normal distribution and $ A_{k+1}$ represents ground truth binary mask for next object $\mathcal{O}_{k+1}$.

\textbf{SmartMask Inference}. 
During inference, given an input image $\mathcal{I}$, object description $\mathcal{T}_{obj}$ and a textual description $\mathcal{T}_{context}$ describing the final scene context,
we first use a panoptic semantic segmentation model \cite{kirillov2019panoptic} to obtain the corresponding semantic layout map $\mathcal{S}_\mathcal{I}$. The generated semantic layout $\mathcal{S}_\mathcal{I}$ is then directly used as input to the above trained diffusion model in order to predict the fine-grained mask $\mathcal{M}_{obj}$ for the target object.
\begin{align}
    \mathcal{M}_{obj} = \mathcal{D}_{\theta} ( \mathcal{E}_{\phi}(\mathcal{S}_I), \mathcal{T}_{obj}, \mathcal{T}_{context}),
\end{align}
where $\mathcal{D}$ refers to the trained diffusion model  from above.

\begin{figure*}[h!]
\vskip -0.2in
\begin{center}
\centerline{\includegraphics[width=1.\linewidth]{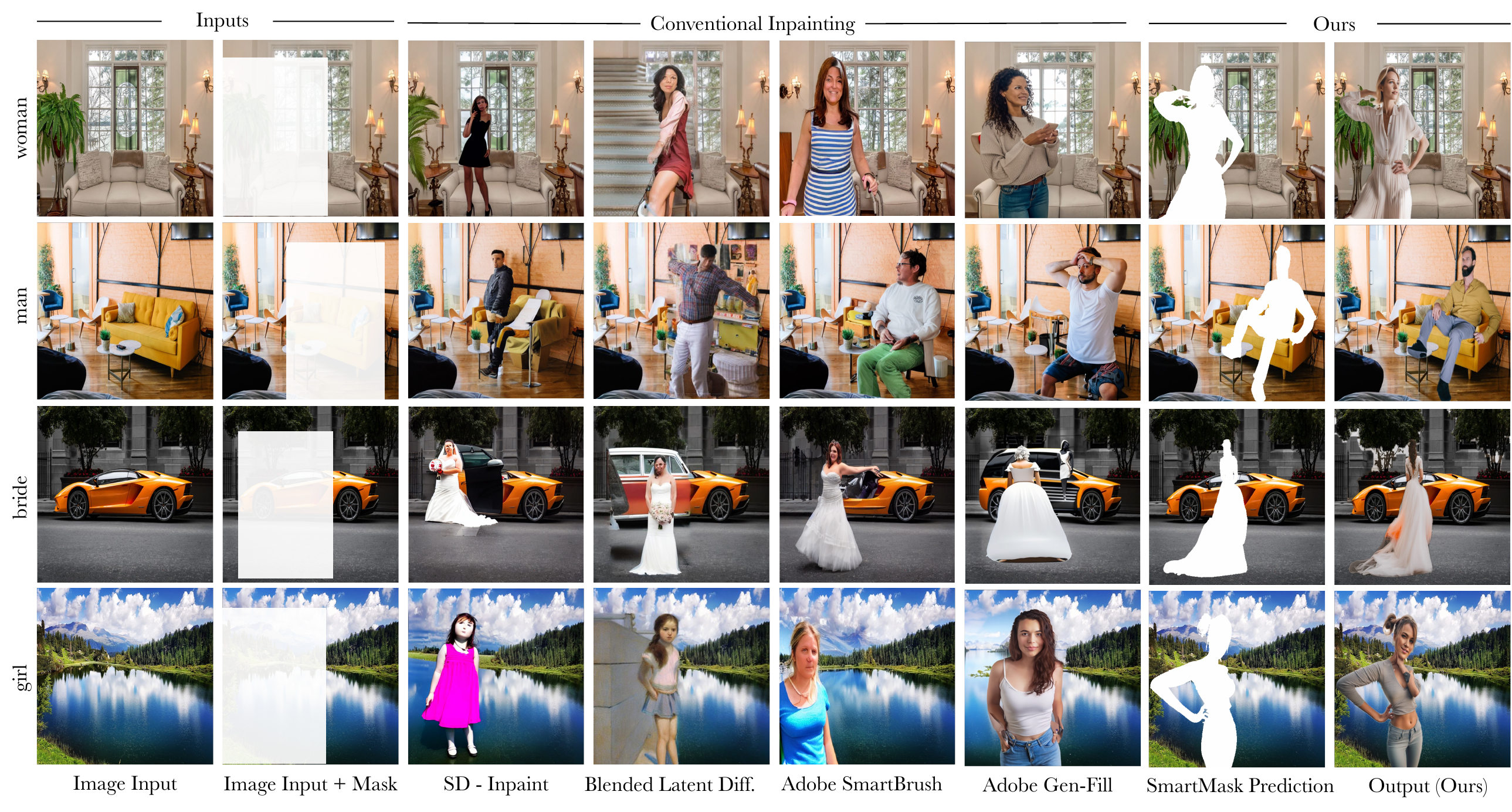}}
\vskip -0.1in
\caption{\emph{\textbf{Qualitative Results for Image Inpainting.}} We observe that as compared to
with state-of-the-art image inpainting \cite{xie2023smartbrush,rombach2021highresolution,adobe2023firefly,avrahami2023blended} methods, \Smask allows the user to perform object insertion while better preserving the background around the inserted object. }
\label{fig:inpainting}
\end{center}
\vskip -0.4in
\end{figure*}

\subsection{Data Augmentation for Precise Mask Control}
\label{sec:data-aug}

While the diffusion model trained in Sec.~\ref{sec:seq-layer-pred}, allows the user to perform mask-free object insertion at diverse positions and scales, the user may also wish to obtain more direct control over the spatial location and details of the inserted object. To this end, we propose a simple train-time data augmentation strategy which allows the user additional control over the output predictions. 
In particular, given the intermediate layer map $S_k$ computed using Eq.~\ref{eq:intermediate-layer} and ground truth mask $A_{k+1}$ for the next object, we randomly replace the input $S_k$ to the diffusion model as,
\begin{align}
    \Tilde{S_k} = g(S_k,G_{obj}) =  S_k \odot (1 - \alpha \ G_{obj})  +  \alpha \  G_{obj}, \label{eq:data-aug}
\end{align}
where $G_{obj}$ is the additional guidance input (\eg, bounding box mask, coarse scribbles \etc) provided by the user and $\alpha=0.7$ helps add additional guidance while still preserving the content of the original input $S_k$ after data augmentation.

\textbf{Training.} In this paper, we mainly consider four main guidance inputs $G_{obj}$ for additional mask control at training time. \emph{1) Mask-free guidance:} in absence of any additional user inputs, we use $G_{obj} = \mathbf{0}^{H,W}$ which prompts the model to suggest fine-grained masks for object insertion at diverse
positions and scales.  \emph{2) Bounding-box guidance:} we set $G_{obj}$ as a binary mask corresponding to ground truth object mask $A_{k+1}$. \emph{3) Coarse Spatial Guidance:} Expecting the user to provide precise bounding box for object insertion is not always convenient and can lead to errors if bounding box is not correct. We therefore introduce a coarse spatial control  where  user may provide a coarse spatial location and the model learns to infer the best placement of the object around the suggested region (Fig.~\ref{fig:mask-controls}c). During training, the same is achieved by setting $G_{obj}$ as a coarse gaussian blob centered at the ground truth object mask $A_{k+1}$.  \emph{4) User scribbles:} Finally, we also allow the user to describe target object using free-form coarse scribbles, by setting $G_{obj}$ as the dilated mask output of ground-truth object mask $A_{k+1}$.


\textbf{Inference.} At inference time, the additional guidance input $G_{obj}$ (\eg bounding box mask, coarse scribbles \etc)  is directly provided by the user.  Given an input image $\mathcal{I}$ with semantic layout $\mathcal{S}_{\mathcal{I}}$, we then use transformation from Eq.~\ref{eq:data-aug} as input to the \Smask diffusion model in order to generate object insertion suggestions with additional control.

\subsection{Global Planning for Multi-Step Inference}
\label{sec:global-planning-model}

While the original \Smask model allows the user to generate fine-grained masks for single object insertion, we would also like to use \Smask for iterative use cases such as multiple object insertion (Sec.~\ref{sec:multi-object-insertion}) or designing a fine-grained layout from scratch with large number ($>10$) of scene elements. Such an iterative use of \Smask would require the model to carefully plan the spatial location of each inserted object to allow for the final scene to be consistent with the final scene context description $\mathcal{T}_{context}$.

To achieve this, we train a visual-instruction tuning \cite{liu2023visual} based planning model which given the input semantic layout $\mathcal{S}_I$, learns to plan the positioning of different scene elements over long sequences. Given an object description $\mathcal{T}_{obj}$ and final scene context  $\mathcal{T}_{context}$, the global planning model provides several bounding box suggestions for object insertion. \Smask model then uses the above predictions as coarse spatial guidance input $G_{obj}$ to provide fine-grained mask suggestions for the next object. The above process can then be repeated in an iterative manner until all objects have been added to the scene (refer Fig.~\ref{fig:layout-generation}a). 

The global planning model is trained in two stages. 1) \textbf{ Feature Alignment.} Typical instruction tuning models are often trained on real images. In contrast, as discussed in Sec.~\ref{sec:seq-layer-pred} we would like to model our problem in semantic space as it allows us to leverage amodal segmentation data for training. To address this domain gap, we first finetune an existing LLaVA model \cite{liu2023visual} to understand the semantic inputs. To do this, given an intermediate semantic map $S_k$ computed using Eq.~\ref{eq:intermediate-layer}, we finetune the projection matrix $\mathbf{W}$ of the LLaVA model \cite{liu2023visual} $\mathcal{H}$ to predict the semantic object labels $\{\mathcal{O}_1, \mathcal{O}_2 \dots \mathcal{O}_k\}$ described in the current scene as,
\begin{align}
    \mathcal{L}_{align} (\mathbf{W}) = \mathcal{L}_{CE} (<\mathcal{O}_1, \mathcal{O}_2 \dots \mathcal{O}_k>, \mathcal{H}(S_k)).
\end{align}

\noindent
2) \textbf{Instruction-Tuning.} Finally, keeping the visual encoder weights for LLaVA model fixed, we next finetune both project matrix $\mathbf{W}$ and LLM weights $\Phi$ \cite{zheng2023judging} for global object planning. In particular, given an intermediate semantic map $\mathcal{S}_k$, we first compute the bounding box coordinates $\mathcal{B}_{k+1} = \{x_{min},y_{min},x_{max},y_{max}\}$ for the next object $\mathcal{T}_{obj} = \mathcal{O}_{k+1}$ using ground truth object mask $\mathcal{A}_{k+1}$. The LLaVA based planning model $\mathcal{H}$ is then trained as,
\begin{align}
    \mathcal{L}_{instruct} (\mathbf{W}, \Phi) = \mathcal{L}_{CE} (\mathcal{B}_{k+1}, \mathcal{H}(S_k, \mathcal{T}_{obj}, C )),
\end{align}
where $C$ represents the caption for the final scene (obtained using ground-truth image $\mathcal{I}$) and provides the model context for placing different scene elements in the image.

\section{Experiments}
\label{sec:experiments}

\textbf{Training Data Collection.} As discussed in Sec.~\ref{sec:method}, we note that a key idea behind \Smask is to model the object insertion problem in semantic space (instead of pixel space), which allows us to leverage semantic amodal segmentation data to obtain large-scale paired training annotations for single or multiple object insertion. However, traditional datasets for semantic amodal segmentations such as COCO-A \cite{zhu2017semantic} (2,500 images, 22,163 instances) and KINS \cite{qi2019amodal} (7,517 images, 92,492 instances) though containing fine-grained amodal segmentation annotations may lack sufficient diversity to generalize across different use-cases.

To address this, we curate a new large-scale dataset
 consisting of fine-grain amodal segmentation masks for different objects in an input image.
The overall dataset consists of 32785 diverse real world images
and a total of 725897 object instances across more than 500 different semantic classes (\eg man, woman, trees, furniture \etc). Each image $\mathcal{I}$ in the dataset consists of a variable number of object instances $\{A_1, A_2, \dots A_n\}, \ n \in [2,50]$ and is annotated with an ordered sequence of semantic amodal segmentation maps $\{S_1, S_2 \dots  S_n\}$. The detailed descriptions $C_I$ for each image are obtained using the InstructBLIP \cite{instructblip} model.

\begin{figure*}[h!]
\vskip -0.2in
\begin{center}
\centering
     \begin{subfigure}[b]{0.495\textwidth}
         \centering
         \includegraphics[width=\textwidth]{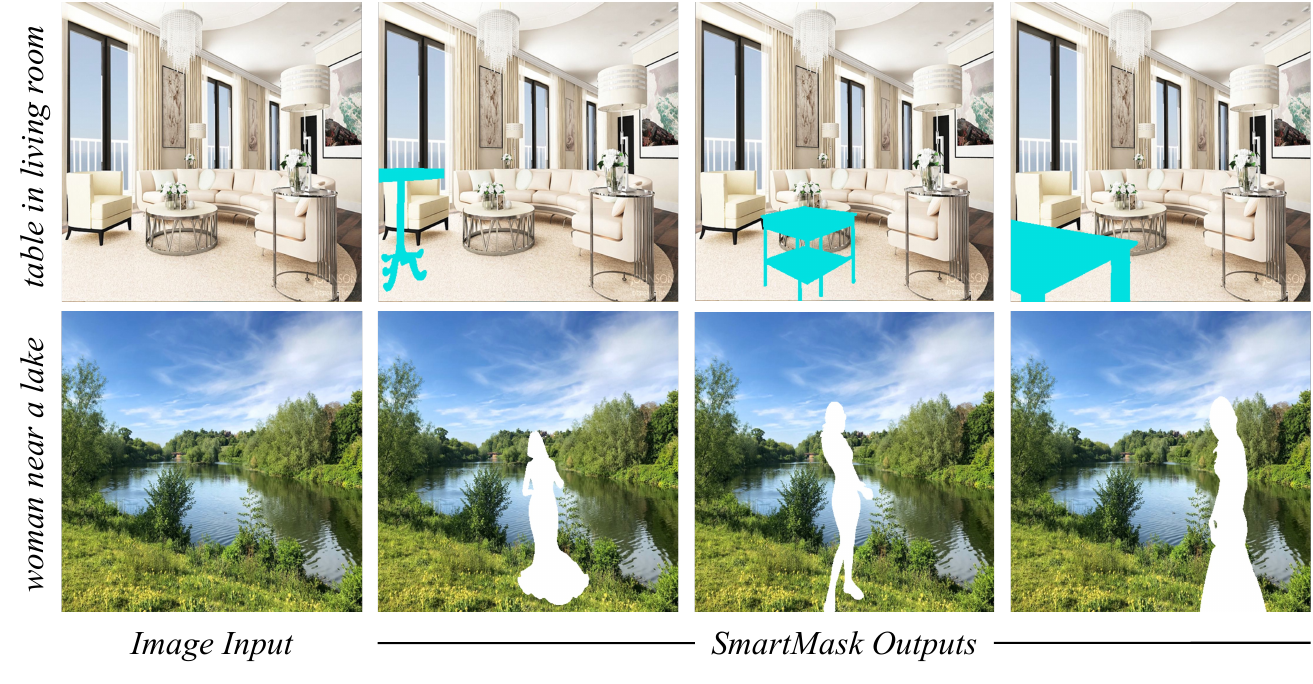}
         \vskip -0.05in
         \caption{\emph{Mask-free object insertion}}
     \end{subfigure}
     \hfill
     \begin{subfigure}[b]{0.495\textwidth}
         \centering
         \includegraphics[width=\textwidth]{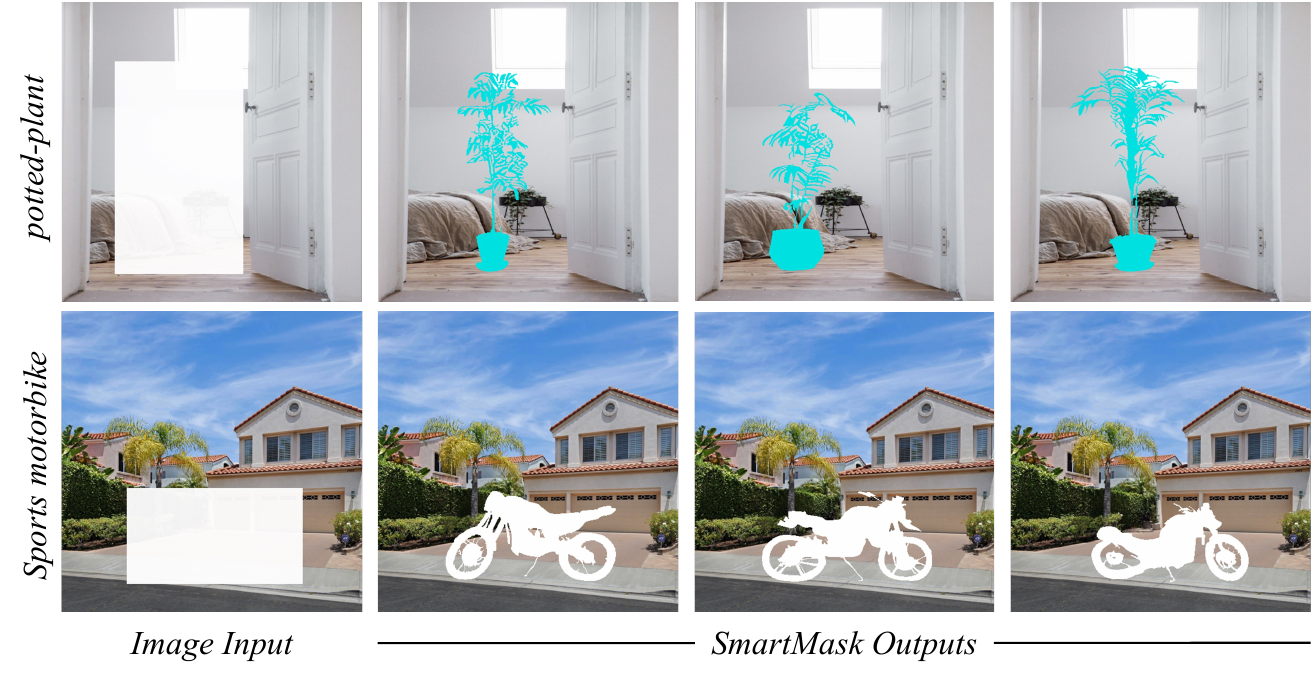}
         \vskip -0.05in
         \caption{\emph{Bounding-box guidance for mask generation}}
     \end{subfigure}
     \vskip -0.04in
     \begin{subfigure}[b]{0.495\textwidth}
         \centering
         \includegraphics[width=\textwidth]{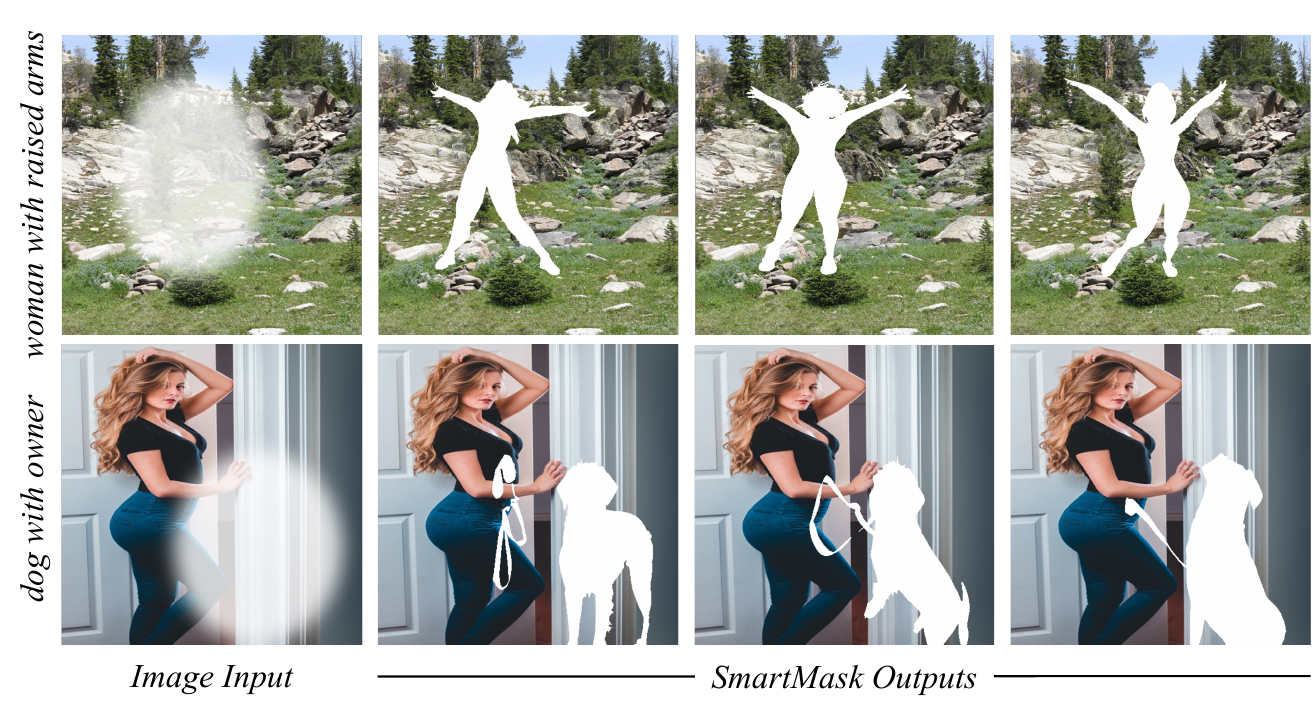}
         \vskip -0.05in
         \caption{\emph{Coarse-spatial guidance for mask generation}}
     \end{subfigure}
     \hfill
    \begin{subfigure}[b]{0.495\textwidth}
         \centering
         \includegraphics[width=\textwidth]{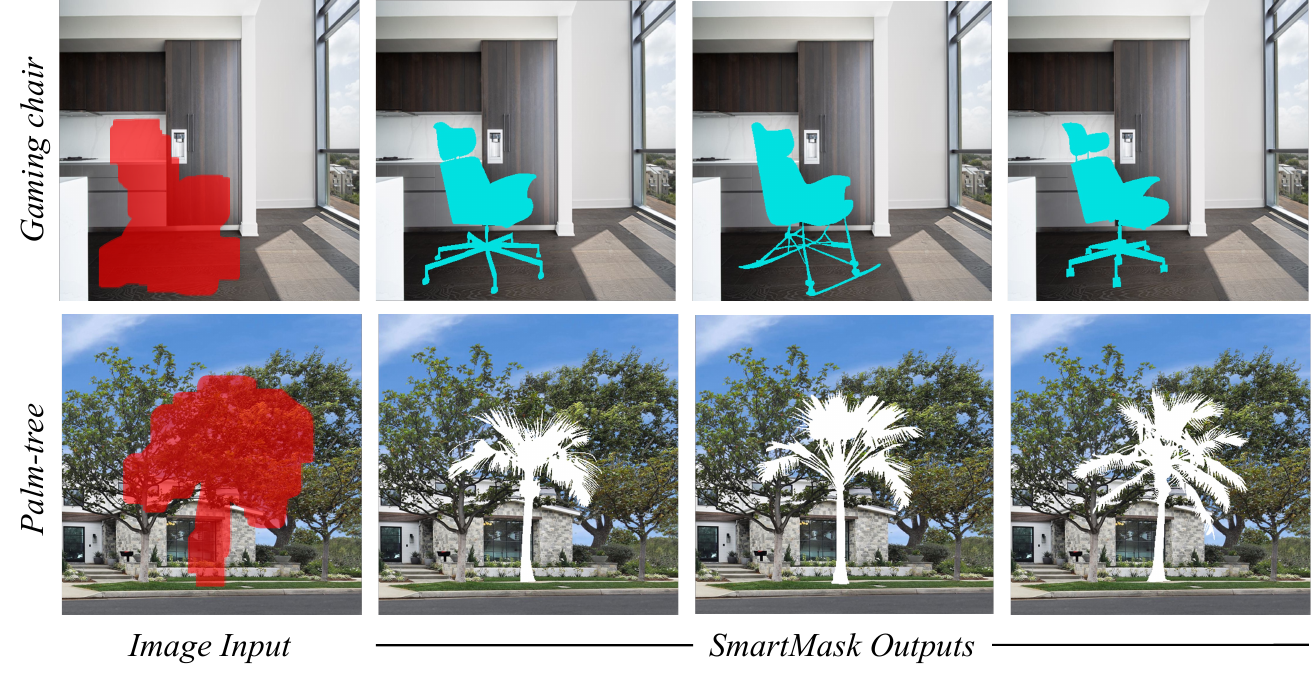}
         \vskip -0.05in
         \caption{\emph{Scribble-based object insertion}}
     \end{subfigure}
\vskip -0.1in
\caption{\textbf{\emph{Diverse User Controls}}. \Smask allows the user to control target-object insertion mask through diverse controls.
}
\vskip -0.3in
\label{fig:mask-controls}
\end{center}
\end{figure*}

\textbf{SmartMask Training.} 
In order to leverage the rich generalizable prior of T2I diffusion models,
we use the weights from publicly available Stable-Diffusion-v1.5 model \cite{rombach2021highresolution} in order to initialize the weights of the \Smask U-Net model trained in Sec.~\ref{sec:seq-layer-pred}. Similar to \cite{brooks2023instructpix2pix}, we modify the architecture of the U-Net model to also condition the output mask predictions on segmentation layout $\mathcal{S}_I$. The \Smask model is trained for a total of 100k iterations with a batch size of 192 and learning rate $1e-5$ using 8 Nvidia-A100 GPUs. During inference, a panoptic semantic segmentation model finetuned on the dataset in Sec.~\ref{sec:experiments} is used for converting real image $\mathcal{I}$  to its semantic layout $\mathcal{S}_I$. ControlNet model trained with SDXL backbone was used to perform precise object insertion with SmartMask outputs.
Please refer the supp. material for further training details.

\subsection{SmartMask for Object Insertion}
\label{sec:inpaint-results}

\emph{\textbf{Baselines.}} We compare the performance of our approach with prior works on performing multi-modal image-inpainting using a textual description and coarse bounding-box mask. In particular we show comparisons with SD-Inpaint \cite{rombach2021highresolution}, SDXL-Inpaint \cite{podell2023sdxl}, Blended-Latent Diffusion \cite{rombach2021highresolution},  and Adobe SmartBrush \cite{xie2023smartbrush}. We also compare the performance of our approach with state-of-the-art commercial inpainting tools by reporting results on recently released Generative-Fill from Adobe Firefly \cite{adobe2023firefly}.

\begingroup
\setlength{\tabcolsep}{4.0pt}
\small
\begin{table}[t]
\begin{center}
\small
\footnotesize
\begin{tabular}{l|ccc}
\toprule
\multirow{2}{*}{Method} & \multicolumn{3}{c}{Evaluation Criteria} \\
\cline{2-4} 
 &  Local-FID $\downarrow$ & CLIP-Score $\uparrow$ & Norm.~L2-BG $\downarrow$ \\
\hline
SD Inpaint \cite{rombach2021highresolution} & 22.31 & 0.249 &   0.374 \\
SDXL Inpaint \cite{podell2023sdxl} & 21.84 & 0.235 &  0.623 \\
Blended L-Diffusion \cite{avrahami2023blended} & 39.77 & 0.253 &  0.451 \\
Adobe SmartBrush \cite{xie2023smartbrush}  & 17.94 & 0.262 &   0.304 \\
Adobe Gen-Fill$^\star$ \cite{adobe2023firefly} & N/A & \textbf{0.268} & 0.289 \\
\hline
Smartmask (Ours) & 19.21 & 0.261 &  \textbf{0.098} \\
\bottomrule
\end{tabular}
\end{center}
\vskip -0.2in
 \caption{\emph{\textbf{Quantitative results for image inpainting}.} We observe that in comparison with state-of-art image inpainting methods, our approach leads to better  preservation of background regions.}
\label{tab:inpainting-quant}
\vskip -0.19in
\end{table}
\endgroup

\emph{\textbf{Evaluation Metrics.}} Following  \cite{xie2023smartbrush}, we report the  results for image inpainting using \emph{1) Local-FID} \cite{heusel2017gans} which measures the realism of the generated objects,  \emph{2) CLIP-Score}  \cite{hessel2021clipscore} which measures the alignment between the textual description and the generated object, and \emph{3) Norm.~L2-BG:} which reports the normalized $\emph{L2}$ difference in the background regions before and after insertion, and helps capture the degree to which the background was preserved.

\emph{\textbf{Qualitative Results.}} Results are shown in Fig.~\ref{fig:inpainting}. We observe that when performing image-inpainting using a coarse bounding box mask, traditional inpainting methods usually lead to a lot of changes in the background regions around the inserted object (\eg living room details in row-1\&2,  mountains in row-4 \etc). Adobe SmartBrush \cite{xie2023smartbrush} which is trained to allow better background preservation, shows better performance, however, still suffers from notable changes to background regions. In contrast, by directly predicting a high-fidelity mask for the target object, the proposed approach allows the user to add new objects on the scene with minimal changes to the background image. Furthermore we observe that target object masks are generated in a scene-aware manner, which helps us add new objects while interacting with already existing ones. 
For instance, when adding \emph{`a man to a couch with table in front'} (Fig.~\ref{fig:inpainting}), prior works typically replace the couch and table to insert the target object (\emph{'man'}). In contrast, \smask places the \emph{`man sitting on the couch with his leg on the table'}, and provides a more natural way for inserting objects in complex scenes.

\begin{figure}[h!]
\begin{center}
\centerline{\includegraphics[width=1.\linewidth]{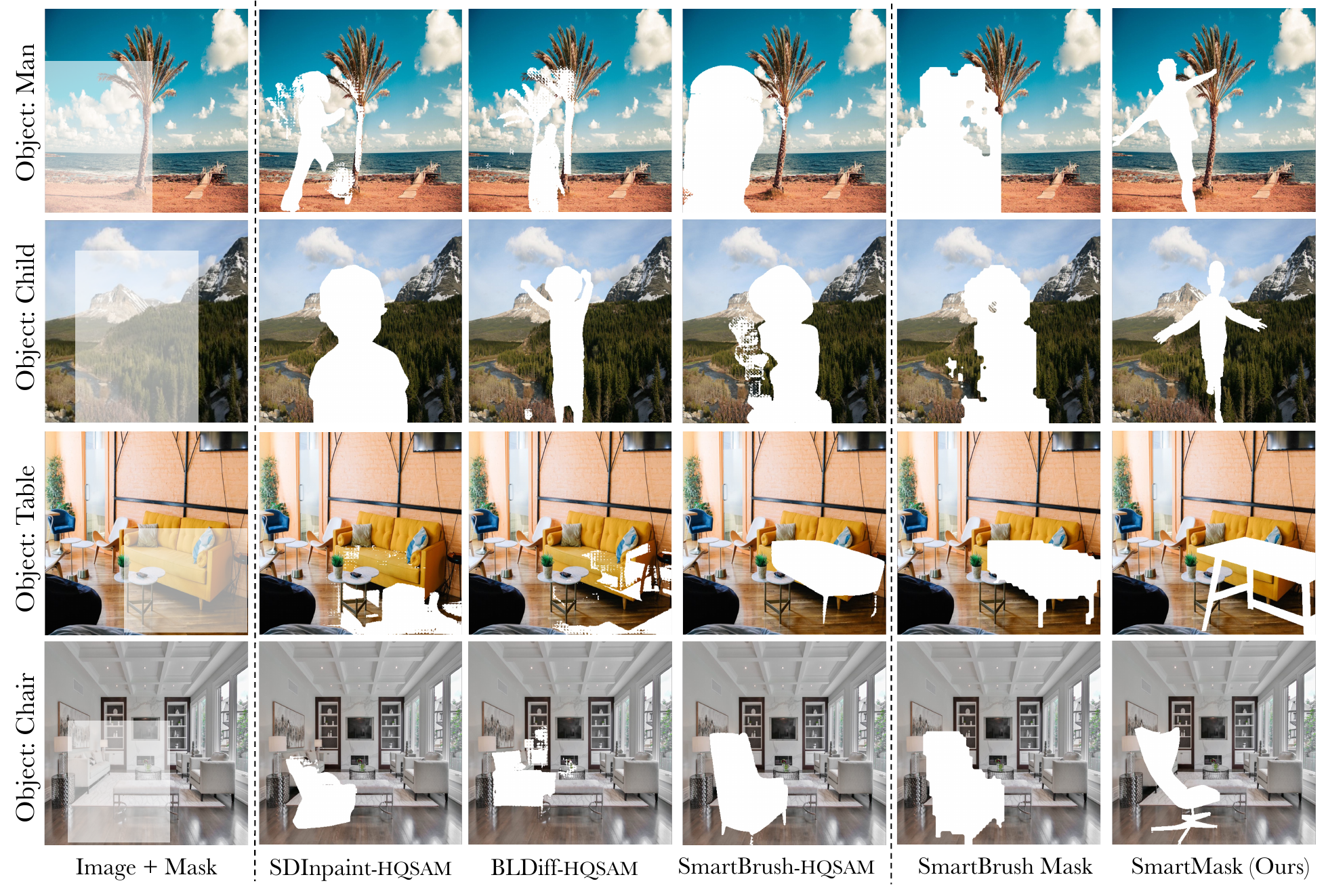}}
\vskip -0.1in
\caption{\emph{\textbf{Comparing output mask quality}} with different Inpaint + HQSAM (middle) methods and SmartBrush mask output (right). }
\label{fig:mask-comp}
\end{center}
\vskip -0.1in
\end{figure}

\begin{figure}[h!]
\vskip -0.15in
\begin{center}
\centerline{\includegraphics[width=1.\linewidth]{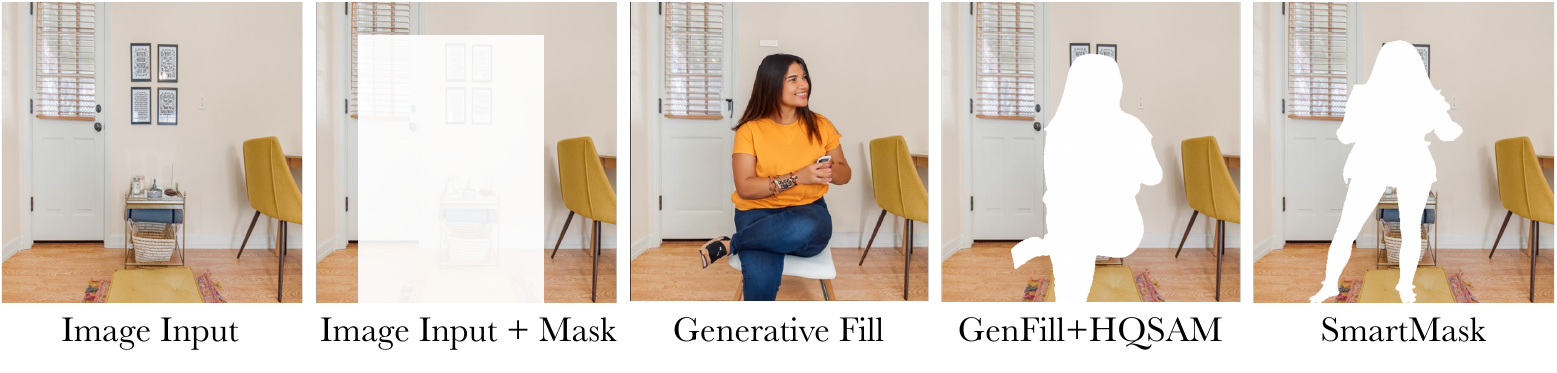}}
\vskip -0.1in
\caption{\emph{\textbf{Limitation of Inpaint + HQSAM}}. In addition to poor mask quality errors (Fig.~\ref{fig:mask-comp}), we observe that \emph{Inpaint+HQSAM} can lead to scene-unaware masks (\eg \emph{mask for woman sitting in air})}.
\label{fig:inpaint-hqsam-limitation}
\end{center}
\vskip -0.4in
\end{figure}

\emph{\textbf{Quantitative Results.}} In addition to qualitative results, we also report the performance of our approach quantitatively 
in Tab.~\ref{tab:inpainting-quant}. We find that
similar to results in Fig.~\ref{fig:inpainting}, 
the proposed approach shows better background preservation (Norm.~L2-BG $\downarrow$) while performing comparably in terms of image quality (local-FID) and text-alignment (CLIP-Score).

\subsection{Evaluating Mask Controllability and Quality}
\label{sec:mask-pred-results}

A key advantage of \Smask is the ability to generate high-quality masks for target object in a controllable manner. In this section, we evaluate the performance of \Smask 
in terms of 1) user control, \& 2) output mask quality.

\textbf{1) User Control}. As shown in  Fig.~\ref{fig:mask-controls}, we observe that \Smask allows the user to control the output object mask in four main ways. \emph{1) Mask-free insertion:} where the model automatically suggests diverse positions and scales for the target object (\eg, \emph{table, woman} in Fig.~\ref{fig:mask-controls}a). \emph{2) Bounding-box guidance:}  which allows user to specify the exact bounding box for object insertion (\emph{potted plant, motorbike} in Fig.~\ref{fig:mask-controls}b). \emph{3) Coarse spatial guidance:} (Fig.~\ref{fig:mask-controls}c) providing a precise bounding box can be challenging for cases with complex object insertions \eg, \emph{dog with owner}. \Smask allows the user to only specify a coarse location for the target object, and the model automatically adjusts the object placement (\ie \emph{dog with head near woman's hand}) to capture object interactions. \emph{3) User-scribbles:} Finally, the user may also control the output shape by providing coarse scribbles. The \smask model can use this as guidance to automatically predict the more finegrain-masks for the target object (\eg, palm-tree, gaming chair in Fig.~\ref{fig:mask-controls}d).

\textbf{2) Output Mask Quality}. We also report results on the quality of generated masks by showing comparisons with the mask-prediction head of the SmartBrush model \cite{xie2023smartbrush}.
Furthermore, we also show comparisons combining standard inpainting methods with HQSAM \cite{sam_hq,kirillov2023segment}. To this end, we first use the provided bounding-box mask to inpaint the target object. The user-provided bounding-box and the inpainted output are then used as input to the HQSAM model \cite{sam_hq} to obtain target object-mask predictions.

Results are shown in Fig.~\ref{fig:mask-comp}. We observe that as compared to outputs of SmartBrush \cite{xie2023smartbrush} mask-prediction head, \Smask generates higher-quality masks with fewer artifacts. Similarly, while using HQSAM on inpainting outputs helps achieve good mask quality for some examples (\eg child in row-2), the HQSAM generated masks (or the inpainted image) often have accompanying artifacts which limits the quality of the output masks. In addition to poor mask quality errors, we also  observe that Inpaint+HQSAM can lead to scene-unaware masks (Fig.~\ref{fig:inpaint-hqsam-limitation}). This occurs because prior inpainting methods typically adds additional objects in the background when performing object insertion. For instance, when inserting \bline{woman in a living room} in Fig.~\ref{fig:inpaint-hqsam-limitation}), we observe that Adobe Gen-Fill \cite{adobe2023firefly} adds an additional chair on which the woman is sitting. 
Extracting only the object mask for such inpainted outputs can lead to scene-unaware masks where \emph{`the woman appears floating in the air`} as the chair was not present in the original image.

The above findings are also reflected in a quantitative user study (Tab.~\ref{tab:mask-quality-quant}), where human subjects are shown a pair of object mask suggestions (ours vs baselines discussed above), and asked to the select the mask suggestion with the higher quality.
As shown in Tab.~\ref{tab:mask-quality-quant}, we observe that \Smask outputs are preferred by majority of human subjects over SmartBrush mask \cite{xie2023smartbrush} and Inpaint + HQSAM outputs.

\begin{figure*}[h!]
\vskip -0.15in
\begin{center}
\centerline{\includegraphics[width=1.\linewidth]{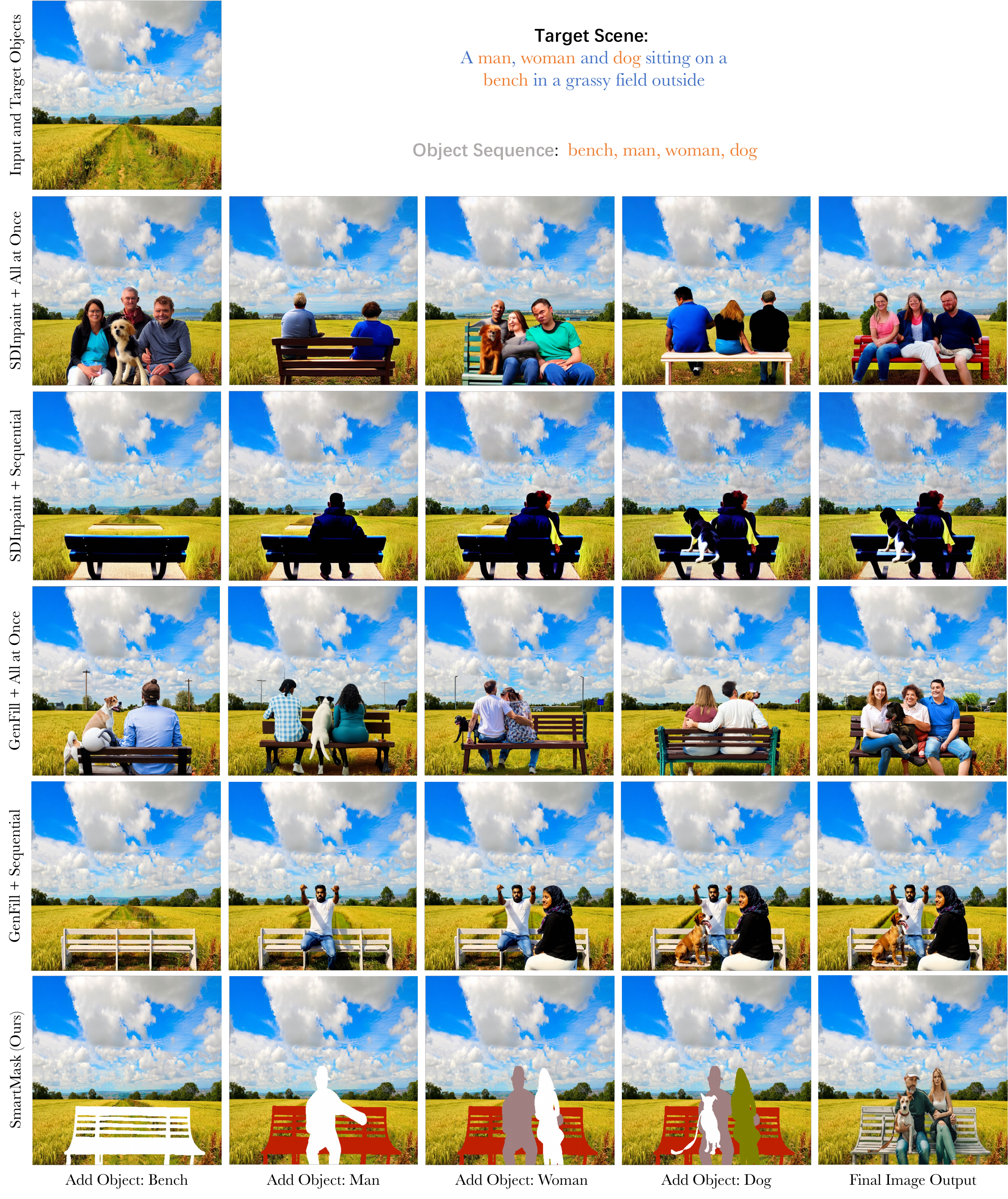}}
\vskip -0.1in
\caption{\emph{\textbf{SmartMask for multi-object insertion.}} We observe that prior state-of-the-art image inpainting methods \cite{rombach2021highresolution,adobe2023firefly} either lead to 1) incorrect objects (\bline{missing dog}) or visual-artifacts (\eg, \bline{people facing bench's back})  when adding all objects at once (row-2, row-4), or, 2) introduce inconsistency-artifacts (\eg, \bline{woman and dog in front of bench} in row-3, \bline{dog and woman not sitting on same bench} in row-5) when adding objects sequentially.  Furthermore, the generated objects (\bline{man, woman and dog} in row-5) can appear non-interacting when generated in a sequential manner.
\Smask helps address this by allowing the user to first add a coherent sequence of context-aware object masks, before using SDXL-based-ControlNet-Inpaint \cite{zhang2023adding,diffusers} model to perform precise object insertion for multiple objects.}
\label{fig:multi-obj-insertion-v1}
\end{center}
\vskip -0.4in
\end{figure*}

\begin{figure*}[h!]
\vskip -0.15in
\begin{center}
\centerline{\includegraphics[width=1.\linewidth]{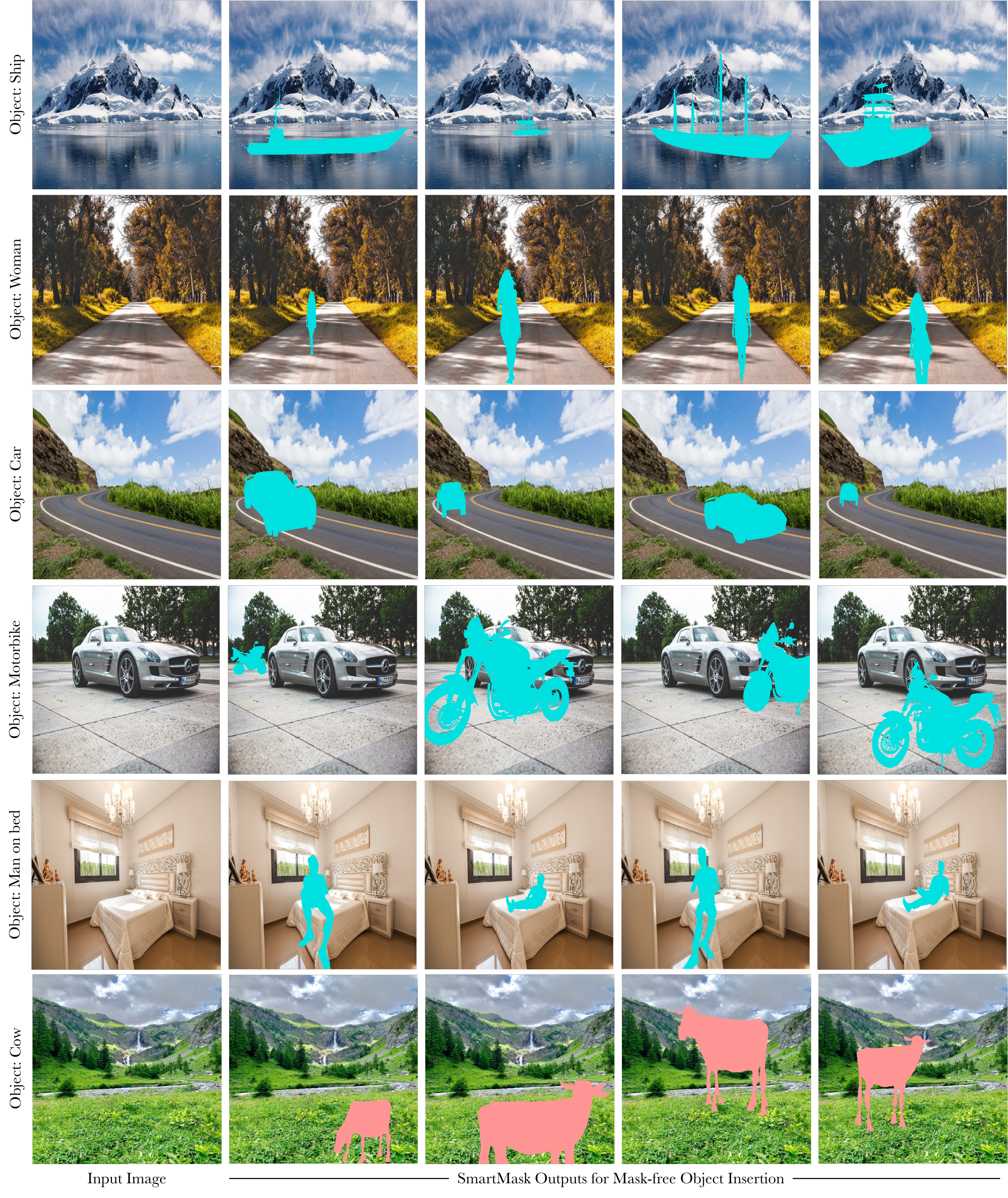}}
\vskip -0.1in
\caption{\emph{\textbf{SmartMask for mask-free object insertion.}} We observe that unlike prior image-inpainting methods which rely on user-provided coarse masks for object location and scale, \Smask also allows for \emph{mask-free object insertion}. This allows the user to generate diverse object insertion suggestions (\eg, \bline{ship}) at different positions and scales. We also observe that the target masks are generated in a scene-aware manner, and can therefore account for the existing scene elements (\eg, \bline{man lying on bed} in row-5, \bline{car riding down the road} in row-3 \etc). We also find that the object insertion suggestions are generated at different scales: thus objects close to camera have larger masks while objects away from camera are predicted using smaller masks (\eg, \bline{motorbike parked beside a car} in row-4).}
\label{fig:mask-free-insertion-v1}
\end{center}
\vskip -0.4in
\end{figure*}

\begingroup
\setlength{\tabcolsep}{7.0pt}
\begin{table}[t]
\begin{center}
\footnotesize
\begin{tabular}{l|ccc}
\toprule
\multirow{2}{*}{Method} & \multicolumn{3}{c}{User Study Results} \\
\cline{2-4} 
 & Win $\uparrow$ & Draw & Lose $\downarrow$ \\
\hline
SDInpaint + HQSAM \cite{rombach2021highresolution,sam_hq}  & 92.04\% & 5.12\% & 2.84\% \\
Blended L-Diff + HQSAM \cite{avrahami2023blended,sam_hq} & 88.91\% & 8.33\% & 2.75\% \\
SmartBrush + HQSAM \cite{xie2023smartbrush,sam_hq} & 63.89\% & 27.78\% & 8.34\% \\
\hline
SmartBrush Mask \cite{xie2023smartbrush} & 91.67\% & 2.78\% & 5.56\%\\
\bottomrule
\end{tabular}
\end{center}
\vskip -0.2in
\caption{\emph{\textbf{User study results}}. For evaluating generated mask quality. We observe that \Smask generates higher quality masks as compared to SmartBrush and various Inpaint+HQSAM methods.}
\label{tab:mask-quality-quant}
\end{table}
\endgroup

\begin{figure*}[h!]
\vskip -0.15in
\begin{center}
\begin{subfigure}[b]{0.95\textwidth}
         \centering
         \includegraphics[width=\textwidth]{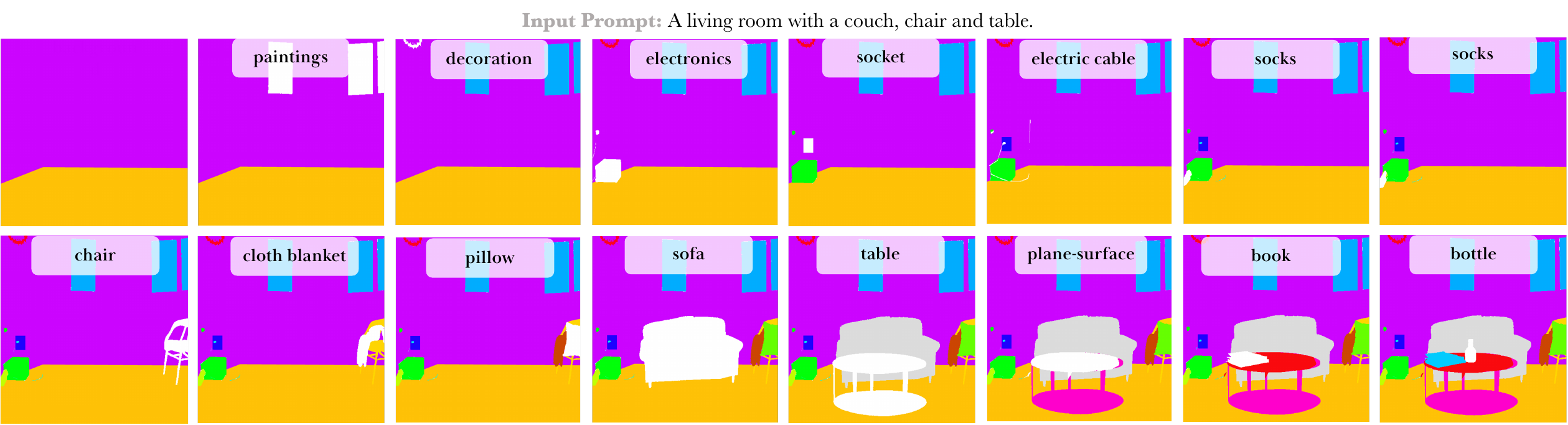}
         \caption{SmartMask for designing very detailed semantic layouts from scratch.}
     \end{subfigure} 
     \begin{subfigure}[b]{0.95\textwidth}
         \centering
         \includegraphics[width=\textwidth]{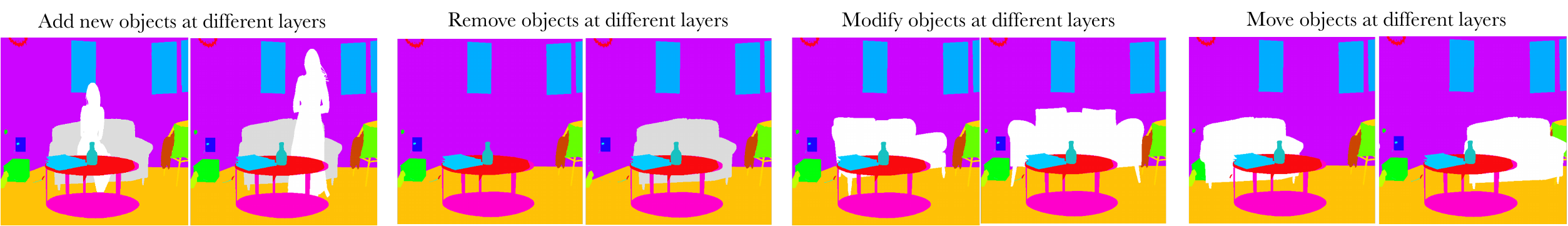}
         \caption{Analyzing controllability of the layouts generated with SmartMask.}
     \end{subfigure}
     \begin{subfigure}[b]{0.95\textwidth}
         \centering
         \includegraphics[width=\textwidth]{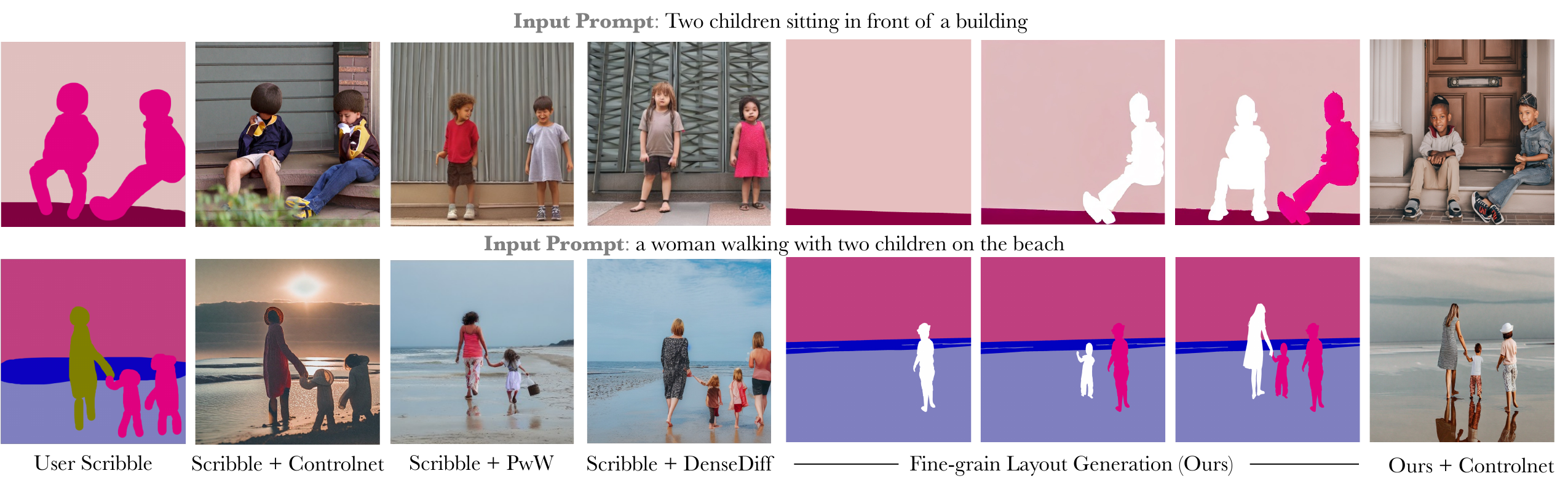}
         \vskip -0.05in
         \caption{Using SmartMask generated layouts for better quality layout-to-image generation.}
     \end{subfigure}
\vskip -0.1in
\caption{\emph{\textbf{Fine-grained layout design.}} We observe that SmartMask when used iteratively, allows the user to generate very detailed layouts from scratch (a). The generated layouts are highly controllable and allow for custom variations through simple layer manipulations (b). }
\label{fig:layout-generation}
\end{center}
\vskip -0.25in
\end{figure*}

\subsection{SmartMask for Multiple Object Insertion}
\label{sec:multi-object-insertion}

While adding a single object to an input image is useful, in practical applications users would typically want to add multiple objects to the input image 
in order to obtain a desired output scene. For instance, given an image depicting a grassy field, the user may wish to add multiple objects \{\bline{bench, man, woman, dog}\} such that the final scene aligns with the context \bline{`a couple with their dog sitting on a bench in a grassy field'} (Fig.~\ref{fig:multi-obj-insertion-v1}). In this section, we show that unlike prior works which are limited to adding each object independently, the proposed approach allows the user to perform multiple-object insertion in a context-aware manner.

Results are shown in Fig.~\ref{fig:multi-obj-insertion-v1}. In particular, we show comparisons with prior inpainting methods when \emph{a)} all objects (\eg bench, man, woman, dog) are inserted all at once, and \emph{b)} different objects are inserted in a sequential manner. We observe that when inserting all objects at once, prior works typically lead to 1) incorrect/missing objects (\eg missing dog, additional person), or, 2) introduce visual-artifacts (\eg, \emph{people facing bench's back}). On the other hand, when adding different objects in sequential manner, we observe that prior works often lead to inconsistency-artifacts. For instance, when adding the \{\bline{bench, man, woman, dog}\} in Fig.~\ref{fig:multi-obj-insertion-v1}, we observe that SD-Inpaint \cite{rombach2021highresolution} leads to outputs which put the \emph{woman and dog on back of the bench}. Commercial state-of-the-art Adobe GenFill \cite{adobe2023firefly} performs better however 
the \emph{dog and woman do not appear to be sitting on the same bench as man}. Furthermore, the generated objects (\bline{man, woman and dog} in row-5) can often appear non-interacting when inserted in a sequential manner. \Smask helps address this by allowing the user to first add a coherent sequence of context-aware object masks, before using SDXL-based-ControlNet-Inpaint \cite{zhang2023adding,diffusers} model to perform precise object insertion.

\begin{figure*}[t]
\begin{center}
\centerline{\includegraphics[width=1.\linewidth]{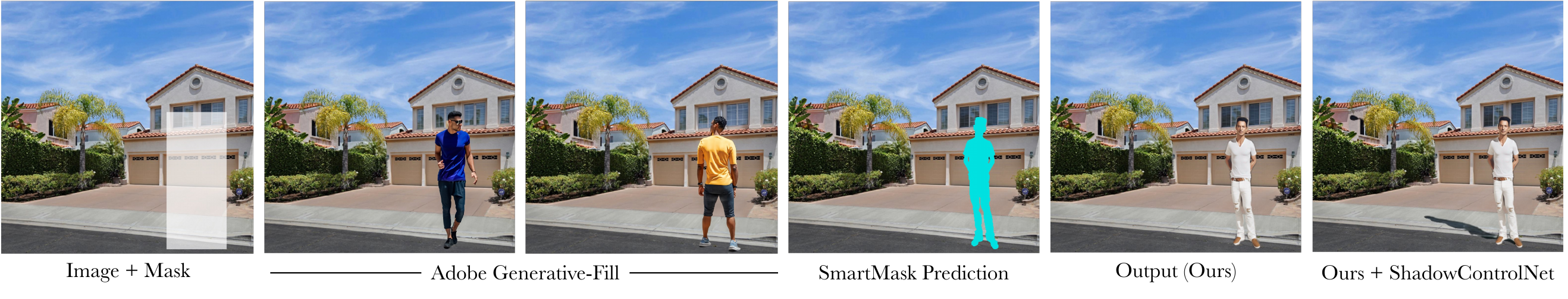}}
\vskip -0.1in
\caption{\emph{Shadow depiction.} We observe that shadow generation remains a challenging problem for even with most state-of-the-art image inpainting methods (\eg, Adobe GenFill \cite{adobe2023firefly} \emph{(left)} and Ours \emph{(right)}). Nevertheless, we find that the \Smask generated precise mask can be used as input to a second shadow-generation ControlNet model \cite{niu2021making} for better quality shadow generation for the inserted object. 
}
\label{fig:lim-shadow-generation}
\end{center}
\vskip -0.3in
\end{figure*}

\subsection{Mask-free Object Insertion}
\label{sec:mask-free object-insertion}
A key advantage of \Smask is that unlike prior works which rely on a user-provided coarse mask, the proposed approach can also be used without user-mask guidance. This allows \Smask to facilitate mask-free object insertion, where it automatically 
provides diverse suggestions for inserting the target-object in the input image.

Results are shown in Fig.~\ref{fig:mask-free-insertion-v1}.  We observe that \Smask also allows for \emph{mask-free object insertion} and enables the user to generate diverse object insertion suggestions for the target object (\eg, \bline{ship} in row-1,  \bline{woman} in row-5) at different positions and scales. We also observe that the masks are generated in a scene-aware manner, and can therefore account for the existing scene elements when adding the new object. For instance, when adding \bline{man on bed} in row-5 Fig.~\ref{fig:mask-free-insertion-v1}, we observe that model autoamtically predicts diverse suggestions where the man is sitting or lying down on the bed in diverse poses. Similarly, when adding \bline{a car riding down the mountain road} in row-3 (Fig.~\ref{fig:mask-free-insertion-v1}), the car is positioned at correct pose and angle on the road region, despite the turning and slanted nature of the road. 



\subsection{SmartMask for Semantic Layout Design}
\label{sec:layout-design-results}

In addition to object insertion, we also find that when used iteratively along with a visual-instruction tuning based planning model from Sec.~\ref{sec:global-planning-model}, \Smask forms a convenient approach for designing detailed semantic layouts with a large number of fine-grain objects (\eg humans, furniture \etc). Results are shown in Fig.~\ref{fig:layout-generation}a. We observe that given a sequence of user provided scene elements (\eg painting, sofa, chair \etc), \Smask generates the entire scene layout from scratch. Furthermore, unlike static layouts generated by a panoptic segmentation model, \Smask generated layouts allow the user greater control over the details of each scene element. Since each object in the final layout is represented by a distinct object mask, the final layouts are highly controllable and allow for a range of custom operations such as adding, removing, modifying or moving objects through simple layer manipulations. (refer Fig.~\ref{fig:layout-generation}b).

\textbf{Controllable S2I Generation.} 
Layout to image generation methods \eg, ControlNet \cite{zhang2023adding}
enable the generation of controllable image outputs from user-scribble based semantic segmentation maps.
However, generating the user-desired layouts with coarse scribbles can itself be quite challenging for scenes with objects that require fine-grain details for best description (\eg humans, chairs \etc). As shown in Fig.~\ref{fig:layout-generation}c, we find that this can lead to image outputs with either deformity artifacts (child in row-1) or incorrect description (woman and children in row-2) when using ControlNet \cite{zhang2023adding}. 
A similar problem is also observed in other coarse-scribble based S2I methods such as DenseDiffusion \cite{densediffusion} and Paint-with-Words (PwW) \cite{balaji2022ediffi}, which provide coarse control over object position but are unable to control finegrain details such as pose, action \etc of the target object.
\Smask helps address this problem by allowing any novice user to generate controllable (Fig.~\ref{fig:layout-generation}b) fine-grain layouts from scratch, which can allow users to better leverage existing S2I methods \cite{zhang2023adding} for higher quality layout-to-image generation (refer Fig.~\ref{fig:layout-generation}c).

\section{Discussion and Limitations}
\label{sec:discussion}

While the proposed approach allows for better quality object insertion and layout control, it still has some limitations. \emph{First,} recall that current \Smask model is trained to predict object insertion suggestions based on the semantic layout $\mathcal{S}_I$ of the input image $\mathcal{I}$. While this 
allows us to leverage large-scale semantic amodal segmentation datasets \cite{qi2019amodal,zhu2017semantic} for obtaining high quality paired annotations during training,
the use of semantic layout input for target mask prediction can be also be limiting as the semantic layout $\mathcal{S}_I$ typically has less depth context as opposed to the original image $\mathcal{I}$. In future, using a ControlNet generated S2I image as pseudo-label can help better train the model to directly predict the target object masks from the original image $\mathcal{I}$.

\emph{Second,} we note that in order to facilitate background preservation and mask-free object insertion, the current \Smask model is trained on a semantic amodal segmentation dataset consisting a total of 32785 diverse real world images (with $\sim 0.75$M different object instances). In contrast, typical object inpainting models such as Adobe SmartBrush \cite{xie2023smartbrush}, SDInpaint \cite{rombach2021highresolution,diffusers} are trained on datasets \cite{schuhmann2021laion} which are orders of magnitude larger in comparison (\eg Adobe SmartBrush \cite{xie2023smartbrush} and SDInpaint \cite{rombach2021highresolution} are trained on 600M samples from the LAION-Aesthetics-v2 5+ dataset \cite{schuhmann2021laion}).  While utilizing the generalizable prior of a pretrained Stable-Diffusion v1.5 model \cite{diffusers} allows our approach to generalize across diverse object categories (\eg mountains, waterfalls, building/towers, humans, dogs, cats, clouds, trees, furniture, electrical appliances \etc) with limited data, generating precise object masks for significantly out-of-distribution objects \eg, dragons remains challenging.  In future, the use of larger training datasets and stronger prior mode (SDXL \cite{podell2023sdxl}) can help alleviate this problem. 

\emph{Finally,} we note that similar to prior inpainting methods \cite{xie2023smartbrush, adobe2023firefly, podell2023sdxl,rombach2021highresolution}, 
accurate shadow-generation around the inserted object remains a challenging problem. For instance, in Fig.~\ref{fig:lim-shadow-generation}, when adding a man in front of a house on a sunny day, we observe that both Adobe GenFill \cite{adobe2023firefly} and \Smask lead to limited shadow depiction around the inserted object. Nevertheless, we find that the \Smask generated precise mask can
be used as input to a second shadow-generation ControlNet model \cite{niu2021making} for better quality shadow generation for the inserted object.
That said, we note that precise shadow generation for the inserted object remains a challenging problem (with both prior work and ours). However, the same is out of scope of this paper, and we leave it
as a direction for future research.

\section{Conclusion}
\label{sec:conclusion}

In this paper, we present \Smask which allows a novice user to generate scene-aware precision masks for object insertion and finegrained layout design.  Existing methods for object insertion typically rely on a coarse bounding box or user-scribble input which can lead to poor background preservation around the inserted object. To address this, we propose a novel diffusion based framework which leverages semantic amodal segmentation data in order to learn to generate fine-grained masks for precise object insertion. When used along with a ControlNet-Inpaint model, we show that the proposed approach achieves superior object-insertion performance, preserving background content more effectively than previous methods. Additionally, we show that 
\Smask provides a highly controllable approach for designing detailed layouts from scratch. As compared with user-scribble based layout design, we observe that the proposed approach can allow users to better leverage existing S2I methods for higher quality layout-to-image generation.

\newpage
{\small
\bibliographystyle{ieee_fullname}
\bibliography{smartmask}
}


\end{document}

%% file: preamble.tex
\usepackage[dvipsnames]{xcolor}
